\documentclass{article}

\PassOptionsToPackage{numbers,sort&compress}{natbib}

 \usepackage[preprint]{neurips_2026}

\usepackage[utf8]{inputenc} 
\usepackage[T1]{fontenc}    
\usepackage{hyperref}       
\usepackage{url}            
\usepackage{booktabs}       
\usepackage{amsfonts}       
\usepackage{nicefrac}       
\usepackage{microtype}      
\usepackage{xcolor}         
\usepackage{amsmath} 
\usepackage{graphicx} 
\usepackage{multirow}
\usepackage{booktabs}
\usepackage{threeparttable}

\title{From Global to Local: Rethinking CLIP Feature Aggregation for Person Re-Identification}

\author{%
  Aotian Zheng$^{1*}$ \quad Winston Sun$^{1*}$ \quad Bahaa Alattar$^1$ \quad Vitaly Ablavsky$^{2,1}$ \quad Jenq-Neng Hwang$^1$ \\[0.5em]
  $^1$Department of Electrical and Computer Engineering, University of Washington, Seattle, WA 98195 \\
  $^2$Applied Physics Laboratory, University of Washington, Seattle, WA 98195 \\[0.3em]
  \texttt{\{azheng, wsun12, balattar, hwang\}@uw.edu} \quad \texttt{vxa@uw.edu} \\[0.3em]
  $^*$Equal contribution
}

\begin{document}

\maketitle

\begin{abstract}
CLIP-based person re-identification (ReID) methods aggregate spatial 
features into a single global \texttt{[CLS]} token optimized for 
image--text alignment rather than spatial selectivity, making 
representations fragile under occlusion and cross-camera variation. 
We propose SAGA-ReID, which reconstructs identity representations by 
aligning intermediate patch tokens with anchor vectors parameterized 
in CLIP's text embedding space — emphasizing spatially stable evidence 
while suppressing corrupted or absent regions, without requiring textual 
descriptions of individual images. Controlled experiments isolate the 
aggregation mechanism under two qualitatively distinct conditions — 
synthetic masking, where identity signal is absent, and realistic human 
distractors, where an overlapping person introduces semantically 
confusing signal — with SAGA's advantage over global pooling growing 
substantially as occlusion increases across both conditions. Benchmark 
evaluations confirm consistent gains over CLIP-ReID across standard and 
occluded settings, with the largest improvements where global pooling is 
most unreliable: up to +10.6 Rank-1 on occluded benchmarks. SAGA's 
aggregation outperforms dedicated sequential patch aggregation on a 
stronger backbone, confirming that structured reconstruction addresses 
a bottleneck that backbone quality and architectural complexity alone 
cannot resolve.  Code available at \url{https://github.com/ipl-uw/Structured-Anchor-Guided-Aggregation-for-ReID}
\end{abstract}    

\section{Introduction}
\label{sec:intro}
Person re-identification (ReID) aims to match images of the same individual 
across disjoint camera views and is a fundamental problem in multi-camera 
perception systems~\cite{zheng2016reid, ye2021deep}. Recent approaches 
increasingly adopt vision--language models (VLMs), particularly 
CLIP~\cite{radford2021clip}, to leverage large-scale image--text pretraining 
for improved generalization. Methods such as CLIP-ReID~\cite{li2023clipreid} 
adapt the image encoder to produce identity-discriminative representations by 
aligning a global \texttt{[CLS]} token with textual prompts during training. 
Despite strong performance, this paradigm exhibits a key structural limitation: 
the deployed model reduces to a standard image encoder whose \texttt{[CLS]} 
attention pattern — optimized for image--text alignment rather than spatial 
selectivity — conflates identity-relevant and corrupted regions under occlusion, 
viewpoint shift, and background clutter, precisely where ReID is hardest.

\begin{figure}[t]
\centering
\includegraphics[width=\linewidth]{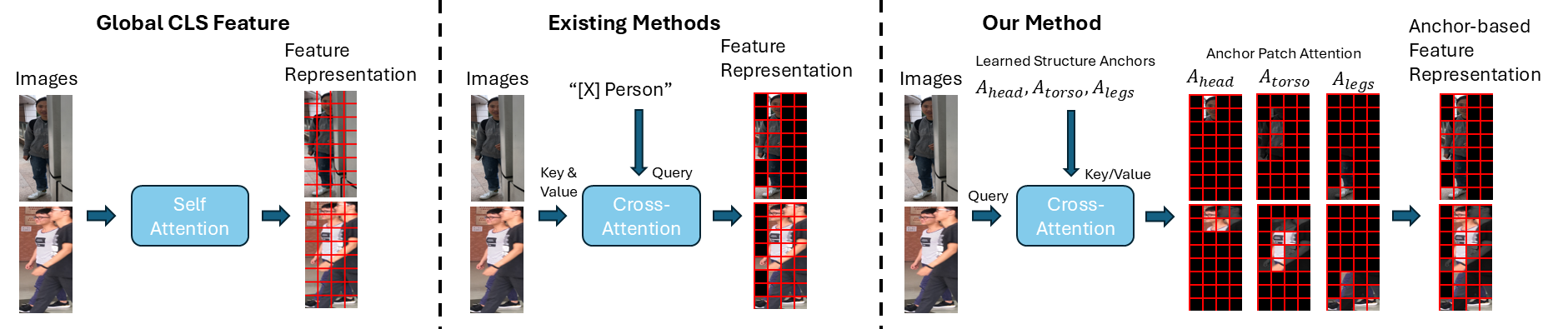}
\caption{Comparison of feature aggregation paradigms for person ReID. 
\textbf{Left:} The global \texttt{[CLS]} token aggregates all patch 
features through self-attention without spatial selectivity. 
\textbf{Middle:} Existing methods use a text prompt as query in 
cross-attention over image patches, suppressing non-person patches — 
effective against background clutter but not human distractor occlusion, 
where target and distractor patches receive comparable attention weights. 
\textbf{Right:} SAGA-ReID uses learned structure anchors as keys and 
values with image patches as queries; target patches align more strongly 
with the anchors than distractor patches do, and the final representation 
is constructed by aggregating anchor-reconstructed patch features rather 
than the original patch features directly.}
\label{fig:methods_comparison}
\end{figure}

\paragraph{Why existing alternatives fall short.}
Part-based methods~\cite{sun2018pcb, he2021transreid} impose spatial structure 
through heuristic partitions or visually learned groupings, but neither approach 
carries a prior about which regions are identity-relevant. More recent CLIP-based 
methods introduce richer language signals or inference-time conditioning. 
CLIP-SCGI~\cite{han2024clipscgi} augments training with LLaVA-synthesized 
captions but discards language at inference. PromptSG~\cite{yang2024pedestrian} 
retains language at inference through a fixed prompt augmented by an 
image-conditioned context token, conditioning visual features on a per-instance 
basis — but this treats language as instance-level descriptive guidance rather 
than a shared structural basis for aggregation. Methods that introduce dedicated 
aggregation mechanisms fare no better without the right inductive bias: 
CLIMB-ReID~\cite{yu2025climb} applies sequential patch filtering via a state 
space model, but the Mamba-specific feature underperforms the global \texttt{[CLS]} 
token on single-image occluded benchmarks, where temporal coherence — the signal 
Mamba relies on — is unavailable. The common failure across all these approaches 
is that none provides a structured basis that governs how patch features are 
\emph{reconstructed} at inference — as opposed to conditioned, filtered, or 
compressed (Figure~\ref{fig:methods_comparison}).

\paragraph{Our approach.}
We propose \textbf{SAGA-ReID} (\textbf{S}tructured \textbf{A}nchor-\textbf{G}uided 
\textbf{A}ggregation for Re-Identification), which uses anchor vectors 
parameterized in CLIP's text embedding space as keys and values in a 
cross-attention module over intermediate patch tokens. Rather than conditioning 
or filtering features, this performs \emph{structured reconstruction}: patch 
features are selectively recombined according to their alignment with a structured 
anchor basis, emphasizing spatially stable evidence while suppressing regions that 
are absent, corrupted, or identity-irrelevant. Text-space initialization provides 
a structured optimization prior that produces more diverse and stable aggregation 
behavior than unconstrained learned bases — not by enforcing explicit semantic 
correspondence, which does not hold at the intermediate layers used for 
aggregation, but by constraining anchors to lie along semantically meaningful 
directions. Shared structured anchors capture structure across images while 
image-conditioned domain anchors provide instance-specific adaptation; at 
inference, the anchor-refined representation is fused with the global 
\texttt{[CLS]} feature.

On Occluded-DukeMTMC, SAGA-ReID achieves 77.8 Rank-1 / 68.3 mAP — a 
+10.6 / +8.0 improvement over CLIP-ReID and a +6.7 / +4.9 improvement over 
the strongest comparable baseline — without external supervision. Gains are 
largest where global pooling is most unreliable. To isolate the aggregation 
mechanism independently of benchmark construction, we conduct controlled 
experiments under two qualitatively distinct occlusion conditions: synthetic 
masking, where identity signal is absent, and realistic human distractors, 
where an overlapping person introduces semantically confusing signal. SAGA's 
advantage over global pooling grows substantially as occlusion increases across 
both conditions, confirming that structured reconstruction addresses a failure 
mode that is not specific to any single occlusion geometry or signal degradation 
type. Critically, SAGA's anchor-guided aggregation outperforms CLIMB-ReID's 
Mamba feature on a stronger backbone, confirming that the benefit is specifically 
attributable to the reconstruction mechanism rather than backbone quality or 
architectural complexity.

\paragraph{Contributions.}
\begin{itemize}
\item \textbf{Identifying the aggregation structure gap.} We identify that 
existing CLIP-based ReID methods share a common limitation regardless of whether 
language is used at training or inference: none provides a structured basis that 
governs how patch features are reconstructed at inference time. Methods that 
condition features on language treat it as instance-level descriptive guidance; 
methods that introduce dedicated aggregation mechanisms lack structured aggregation 
priors. We show this gap is the primary source of fragility under occlusion and 
cross-camera variation.

\item \textbf{Structured anchor-guided reconstruction.} We propose 
SAGA-ReID, which inverts the standard cross-attention polarity used 
in vision-language models: rather than using text as query to 
condition or filter image patches — which cannot distinguish target 
from distractor when both are person patches — image patches attend 
to text-space initialized anchors as keys and values. This enables 
structured feature reconstruction without explicit part labels or 
textual supervision, and to our knowledge represents the first 
application of this inverted cross-attention design to single-image 
ReID aggregation.

\item \textbf{Consistent gains under challenging conditions.} SAGA-ReID achieves 
consistent improvements over CLIP-ReID across standard benchmarks with 
substantially larger gains under occlusion, and outperforms dedicated aggregation 
mechanisms on stronger backbones. Controlled experiments under synthetic masking 
and realistic human distractor occlusion confirm that the advantage grows with 
occlusion severity and generalizes across qualitatively distinct failure modes — 
demonstrating that structured reconstruction addresses a bottleneck that 
conditioning, filtering, and architectural complexity alone cannot resolve.
\end{itemize}

\section{Related Work}
\paragraph{Person re-identification.}
Early ReID methods focused on supervised metric learning using convolutional 
backbones~\cite{hermans2017triplet}. PCB~\cite{sun2018pcb} introduced part-based 
pooling to impose spatial structure, and TransReID~\cite{he2021transreid} extended 
this to transformer architectures with side-information embeddings and jigsaw-patch 
augmentation. Pose-guided~\cite{miao2019pose} and attribute-aware~\cite{zhu2021part} 
methods further improve spatial selectivity under occlusion through explicit keypoint 
or parsing supervision. These approaches operate entirely within the visual feature 
space: spatial groupings are determined by visual attention or predefined partitions, 
with no prior over which regions are identity-relevant. SAGA-ReID addresses this gap 
not through external supervision but by grounding the aggregation process in CLIP's 
text embedding space.

\paragraph{CLIP-based ReID.}
CLIP-ReID~\cite{li2023clipreid} establishes the current paradigm by aligning a 
global \texttt{[CLS]} token with identity-conditioned text prompts, yielding 
substantial gains over ImageNet-pretrained baselines. Subsequent work extends this 
framework to multi-modal and multi-platform settings~\cite{ha2025multimodal}. 
CLIP-SCGI~\cite{han2024clipscgi} augments training with LLaVA-synthesized captions 
to provide fine-grained textual supervision but discards language at inference. 
PromptSG~\cite{yang2024pedestrian} retains language at inference through a fixed 
prompt augmented by an image-conditioned context token, conditioning visual features 
on a per-instance basis. CLIMB-ReID~\cite{yu2025climb} incorporates temporal modeling 
through a hybrid Mamba architecture, achieving strong results on standard benchmarks 
— though as our analysis shows, the Mamba-specific feature underperforms the global 
\texttt{[CLS]} token on single-image occluded benchmarks, suggesting the gains 
reflect improved backbone finetuning rather than effective sequential aggregation. 
Despite their diversity, these methods share a common limitation: none provides a 
structured basis that governs how patch features are reconstructed at inference — 
representations are still derived primarily from a global \texttt{[CLS]} token or 
conditioned features rather than through principled spatial reconstruction. This 
limitation is distinct from simple signal absence: as our controlled distractor 
experiments show, global pooling degrades under semantically confusing occlusion — 
where an overlapping person introduces misleading rather than merely absent signal — 
a condition none of these methods are designed to address.

\paragraph{Feature aggregation and anchor-based methods.}
Cross-attention with learned latent queries — as in 
Perceiver~\cite{jaegle2021perceiver} and slot 
attention~\cite{locatello2020object} — compresses spatial inputs by having 
unconstrained latent vectors summarize patch features. This is structurally 
different from SAGA's approach: rather than compressing features into learned 
slots, SAGA uses anchors as keys and values that patch tokens attend to, 
reconstructing the representation through alignment with a structured basis. 
PromptSG~\cite{yang2024pedestrian} and CLIP-SCGI~\cite{han2024clipscgi} use 
language to condition or supervise visual features, but treat language as 
instance-level descriptive guidance: text conditions feature extraction for 
each image individually. The key distinction from SAGA-ReID is not whether 
language is used, but what role it plays — in SAGA, CLIP's text embedding 
space serves as a fixed structural basis for feature reconstruction rather 
than a per-instance conditioning signal, allowing the aggregation structure 
itself to be grounded in a structured prior without requiring textual 
descriptions of individual images.

\section{Method}
\label{sec:method}
\begin{figure}[t]
\centering
\includegraphics[width=\linewidth]{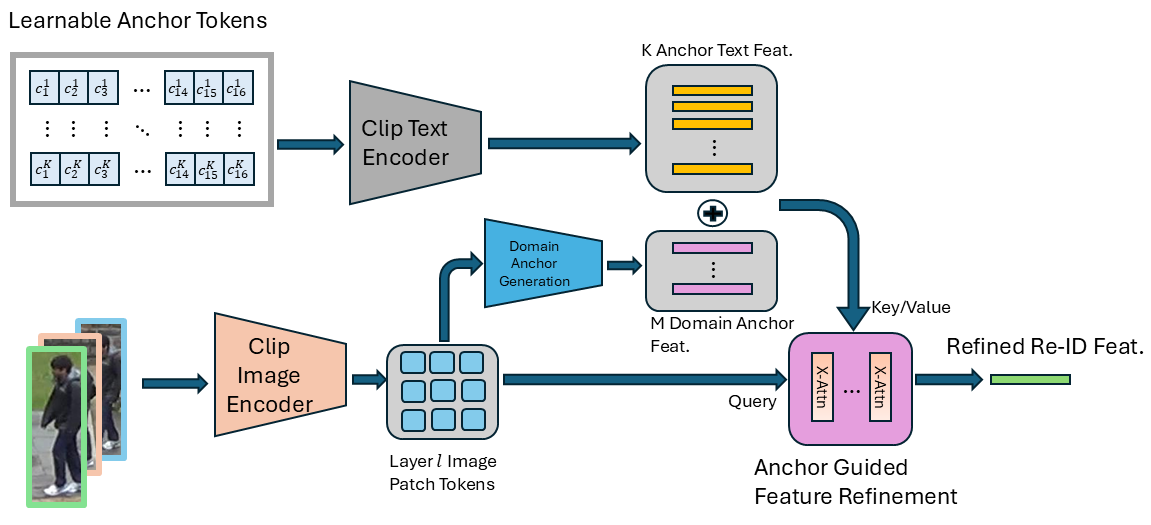}
\caption{SAGA-ReID architecture overview}
\label{fig:method}
\end{figure}

We propose a structured anchor-guided feature aggregation framework for 
CLIP-based person re-identification. Rather than relying on a single global 
\texttt{[CLS]} token, our approach reconstructs a spatially selective 
representation by aligning intermediate patch tokens with a structured anchor 
basis parameterized in CLIP's text embedding space. Our method consists of 
three key components: (1) text-grounded structured anchors that provide a 
structured prior, (2) an image-conditioned domain anchor module for 
instance-specific adaptation, and (3) a cross-attention-based refinement 
module that performs anchor-guided reconstruction over patch tokens. The 
overall pipeline is illustrated in Fig.~\ref{fig:method}.

\subsection{Preliminaries}

Let $\mathcal{I}$ denote a person image. A CLIP ViT-B/16 image encoder $\phi$ 
processes $\mathcal{I}$ through $L$ transformer blocks, producing patch token 
representations $\mathbf{T}^{(\ell)} \in \mathbb{R}^{N \times D}$ at each 
layer $\ell$, where $N$ is the number of patches and $D=768$. The final block 
additionally produces a global \texttt{[CLS]} token $\mathbf{v} \in \mathbb{R}^D$ 
and a projected embedding $\mathbf{p} = \phi_{\text{proj}}(\mathbf{v}) \in 
\mathbb{R}^{512}$ aligned with CLIP's joint embedding space. Standard CLIP-ReID 
methods use $\mathbf{v}$ as the sole image representation; we instead extract 
intermediate patch tokens $\mathbf{T} = \mathbf{T}^{(\ell^*)}$ from layer 
$\ell^* < L$ and perform anchor-guided reconstruction over them.

\subsection{Text-Grounded Anchors}

\paragraph{Motivation.}
The \texttt{[CLS]} attention pattern is optimized for image--text alignment 
rather than spatial selectivity, leaving patch aggregation without a principled 
basis for distinguishing stable identity-relevant regions from corrupted or 
background ones. Unconstrained aggregation mechanisms can learn selective pooling 
from training data but may not generalize under distribution shift. We instead 
parameterize the aggregation basis in CLIP's text embedding space for two 
complementary reasons. First, it provides a structured optimization landscape 
encouraging diverse, semantically meaningful anchor directions — as we show in 
Section~\ref{sec:ablation}, this benefit operates through \emph{optimization 
structure} rather than explicit semantic correspondence, consistent with 
intermediate patch tokens having diverged from CLIP's joint embedding space at 
layer $\ell^*$. Second, text-space initialization combined with the decorrelation 
loss encourages anchors to partition into spatially distinct behaviors — as 
visualized in Section~\ref{sec:visualization} — producing a reconstructed 
feature that captures complementary spatial evidence to the global 
\texttt{[CLS]} token rather than redundantly recapitulating it. Free anchors, 
lacking this structured initialization, tend toward capturing globally salient 
regions already emphasized by \texttt{[CLS]}, limiting the informational 
benefit of fusion. This is consistent with the larger fusion gains observed 
with structured anchors versus free anchors (Section~\ref{sec:ablation}), 
and with the reduced fusion gain when applying SAGA to CLIMB-ReID, whose 
backbone finetuning strategy produces a \texttt{[CLS]} token that already 
captures more locally discriminative evidence, leaving less complementary 
signal for the reconstructed feature to contribute.

\paragraph{Anchor construction.}
We introduce $K$ learnable anchor vectors $\mathbf{A} = [\mathbf{a}_1, \ldots, 
\mathbf{a}_K] \in \mathbb{R}^{K \times D}$ as the structured basis for feature 
reconstruction. Each anchor $\mathbf{a}_k$ is obtained by passing a learnable 
context sequence $\mathbf{c}_k \in \mathbb{R}^{L_c \times 512}$ through the 
frozen CLIP text encoder $\psi$, followed by a linear projection:
\begin{equation}
\mathbf{a}_k = \mathbf{W}_{\text{proj}} \cdot \psi([\mathbf{e}_{\text{sos}}, 
\mathbf{c}_k, \mathbf{e}_{\text{suf}}])
\end{equation}
Context vectors $\{\mathbf{c}_k\}$ are initialised from the CLIP-ReID prompt 
and optimised during training; $\psi$ remains frozen. To prevent anchor 
collapse, we apply a cosine decorrelation loss:
\begin{equation}
\mathcal{L}_{\text{div}} = \frac{\lambda_{\text{div}}}{K(K-1)} \sum_{i \neq j} 
(\hat{\mathbf{a}}_i^\top \hat{\mathbf{a}}_j)^2
\end{equation}
where $\hat{\mathbf{a}}_k = \mathbf{a}_k / \|\mathbf{a}_k\|$ are 
$\ell_2$-normalised anchors.

\subsection{Domain Anchors}

Structured anchors are static across images. To complement them, $M$ domain 
anchors $\mathbf{D} \in \mathbb{R}^{M \times D}$ are computed per image by 
mean-pooling patch tokens and decoding through a two-layer MLP:
\begin{equation}
\mathbf{D} = \text{reshape}\!\left(W_2\,\sigma\!\left(W_1 
\bar{\mathbf{t}}\right)\right) \in \mathbb{R}^{M \times D}
\end{equation}
capturing instance-specific appearance cues such as dominant colour or texture. 
The full anchor set used as keys and values in refinement is $\mathbf{A}^{+} = 
[\mathbf{A};\,\mathbf{D}] \in \mathbb{R}^{(K+M) \times D}$. Only the context 
vectors and domain anchor module are learned; $\psi$ remains frozen.

\subsection{Structured Refinement Module}

Given patch tokens $\mathbf{T}$ and anchors $\mathbf{A}^+$, we apply $n$ 
stacked cross-attention blocks where patch tokens attend to the anchor basis:
\begin{align}
\mathbf{T}^{(i)} &= \mathbf{T}^{(i-1)} + \text{MHA}(\text{LN}
(\mathbf{T}^{(i-1)}), \text{LN}(\mathbf{A}^+)) \\
\mathbf{T}^{(i)} &= \mathbf{T}^{(i)} + \text{FFN}(\text{LN}(\mathbf{T}^{(i)}))
\end{align}
The final attention matrix $\mathbf{W} \in \mathbb{R}^{N \times (K+M)}$ 
reflects how strongly each patch aligns with the anchor basis. We pool by 
taking the maximum alignment across anchors and normalizing:
\begin{equation}
w_n = \max_k \mathbf{W}_{n,k}, \quad \tilde{w}_n = \frac{w_n}{\sum_{n'} w_{n'}}
\end{equation}
The reconstructed feature $\mathbf{f}_{\text{ref}} = \sum_{n} \tilde{w}_n 
\cdot \mathbf{T}^{(n)}_{\text{refined}}$ is projected and batch-normalised to 
produce the final identity embedding. Cross-attention blocks are initialised 
from CLIP layer $\ell^*$.

\subsection{Training Objective}

The model is trained with label-smoothed cross-entropy $\mathcal{L}_{\text{id}}$, 
batch-hard triplet loss $\mathcal{L}_{\text{tri}}$, and an image-to-text 
alignment loss on the projected CLIP feature $\mathbf{p}$:
\begin{equation}
\mathcal{L} = \mathcal{L}_{\text{id}}(\mathbf{f}) + \lambda_{\text{tri}}
\mathcal{L}_{\text{tri}}(\mathbf{f}) + \lambda_{\text{i2t}}\mathcal{L}_{\text{id}}
(\mathbf{p}, \mathbf{T}_{\text{text}}^\top) + \mathcal{L}_{\text{div}}
\end{equation}
The projection head $\mathbf{p}$ serves as an auxiliary supervision branch that 
preserves CLIP's semantic structure without perturbing the backbone; it does not 
improve retrieval at inference and is excluded from fusion ($w_p = 0$).

\subsection{Training Strategy}

Training proceeds in two stages: CLIP-ReID initialisation, anchor and 
refinement module training with frozen backbone. 
This staged optimisation stabilises training and preserves the pretrained 
representation during early learning.

\subsection{Score-Level Fusion}

At inference, refined and backbone features are combined via weighted cosine 
similarity:
\begin{equation}
\mathbf{f}_{\text{fuse}} = \ell_2\text{-norm}\left([\sqrt{w_r}\,
\mathbf{f}_{\text{ref}};\ \sqrt{w_i}\,\mathbf{v}]\right)
\end{equation}
The refined feature carries the highest weight ($w_r \gg w_i$), with the 
backbone \texttt{[CLS]} providing complementary global signal.

\section{Experiments}
If feature aggregation is the primary bottleneck, the following 
predictions follow directly. First, SAGA's advantage over global 
pooling should be negligible at low occlusion where global pooling 
remains effective, grow substantially at intermediate coverage where 
partial identity evidence remains but global pooling increasingly 
conflates informative and corrupted regions, and decline at severe 
occlusion where identity signal is exhausted for any method. Second, 
this peak-then-decline pattern should hold across qualitatively 
distinct occlusion types — both synthetic masking and human distractor 
occlusion — since the bottleneck is in aggregation structure rather 
than any specific signal degradation type. Third, the crossover point 
should shift to higher coverage under human distractor occlusion 
relative to masking, since person patches are harder to suppress 
through anchor alignment than absent patches. Fourth, methods with 
dedicated aggregation mechanisms but no structured prior — specifically 
CLIMB-ReID's Mamba feature — should not show the same pattern. 
Sections~\ref{sec:occlusion}--\ref{sec:feature_analysis} evaluate 
these predictions in order.

\subsection{Robustness under Controlled Masking}
\label{sec:occlusion}
\begin{figure}[h]
\centering
\includegraphics[width=\linewidth]{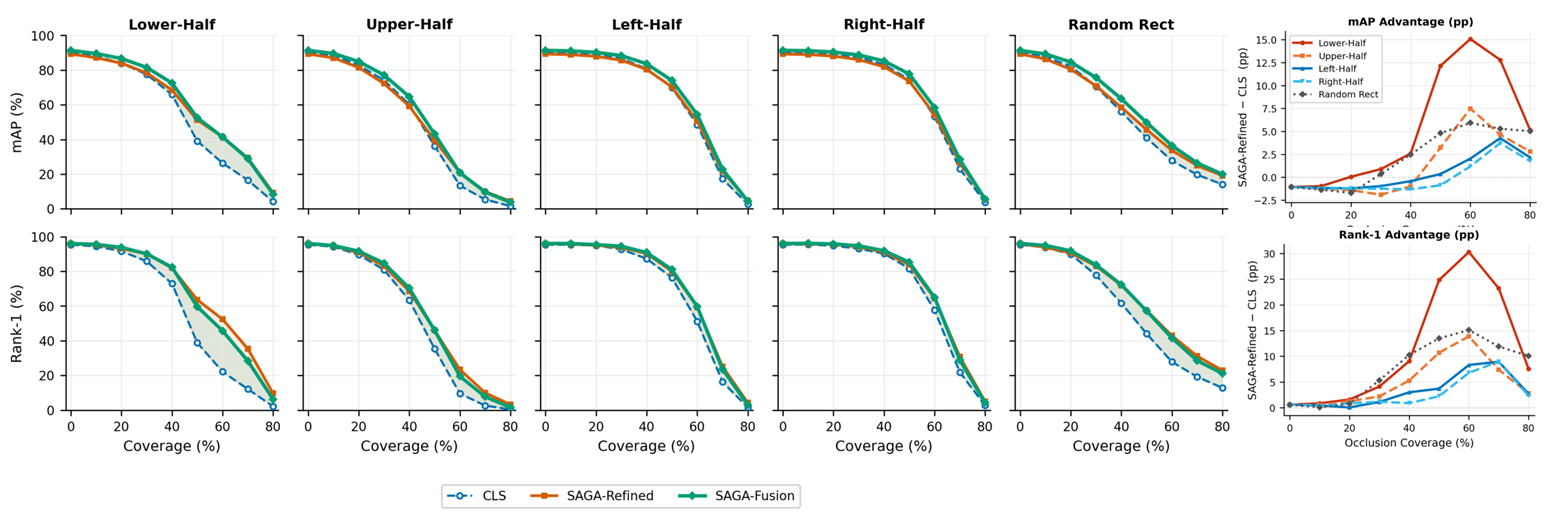}
\caption{Robustness under controlled occlusion on Market-1501. 
Left: mAP and Rank-1 under five occlusion types at increasing 
coverage. Right: SAGA-Refined advantage over \texttt{[CLS]} (pp). 
The advantage peaks at 60-70\% coverage, reaching +30 Rank-1 pp 
and +15 mAP pp under lower-half occlusion, and is consistent 
across all three occlusion geometries.}
\label{fig:occlusion_robustness}
\end{figure}

To directly validate the aggregation mechanism independently of 
benchmark construction, we evaluate on Market-1501 with 
synthetically occluded query images, comparing the global 
\texttt{[CLS]} feature against SAGA-Refined and SAGA-Fusion 
across three occlusion types — lower-half, upper-half, and 
random rectangle — at coverage levels from 0\% to 80\%. Gallery 
images remain unoccluded throughout, reflecting the typical 
deployment scenario where query images are occluded but reference 
gallery images are not. This experiment isolates the case where 
occlusion removes identity signal; Section~\ref{sec:distractor} 
examines the complementary case where an overlapping person 
introduces semantically confusing signal.

Figure~\ref{fig:occlusion_robustness} shows that SAGA maintains 
substantially higher performance than \texttt{[CLS]} as occlusion 
increases, with the advantage growing as coverage increases up to 
40--60\% before declining as occlusion becomes severe enough that 
little identity signal remains for any method. This peak-then-decline 
pattern confirms the first prediction: structured aggregation is 
most beneficial at intermediate occlusion levels where partial 
identity evidence remains but global pooling increasingly conflates 
informative and corrupted regions.

The advantage is largest under lower-half occlusion, reaching 
\textbf{+30 Rank-1 pp} and \textbf{+15 mAP pp} at 60\% coverage. 
This asymmetry reflects the greater discriminative value of 
upper-body cues — clothing color, texture, and style — relative 
to lower-body regions: when the lower body is absent, SAGA's 
structured aggregation concentrates on the remaining upper-body 
signal more effectively than \texttt{[CLS]} pooling, which cannot 
selectively weight spatially available evidence. Upper-half and 
random rectangle occlusion show smaller but consistent advantages, 
confirming the robustness benefit generalizes across occlusion 
geometries rather than being specific to a single spatial pattern.

Notably, SAGA-Refined and SAGA-Fusion track closely across all 
conditions, indicating that the refined feature itself carries 
the robustness benefit and the global \texttt{[CLS]} fusion 
provides modest additional gain rather than being the primary 
source of improvement. Section~\ref{sec:distractor} tests whether 
this advantage extends to a qualitatively harder condition where 
the occluding region is a semantically similar person rather than 
a blank mask.

\subsection{Robustness under Human Distractor Occlusion}
\label{sec:distractor}

\begin{figure}[h]
\centering
\includegraphics[width=\linewidth]{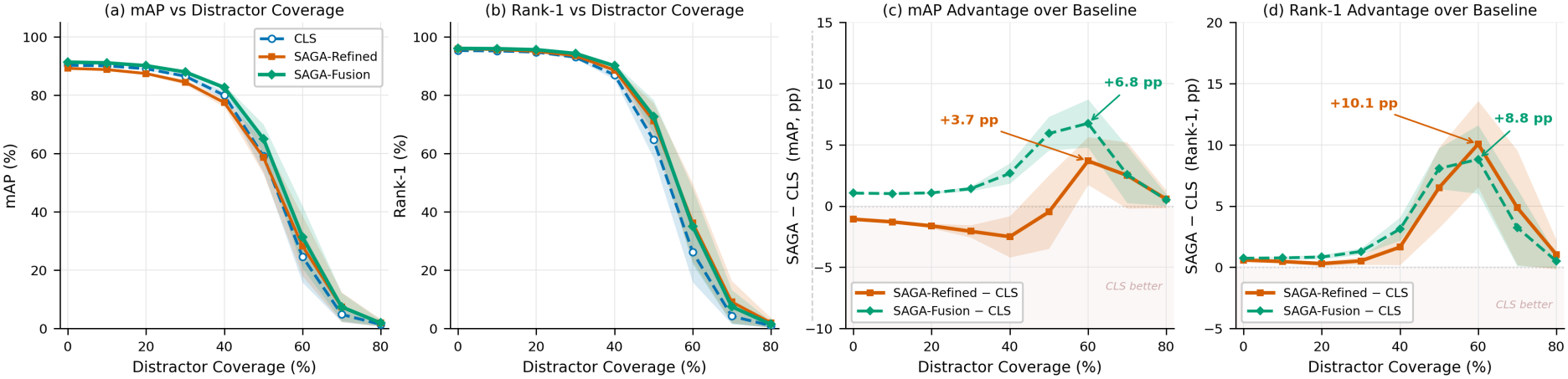}
\caption{Robustness under human distractor occlusion on Market-1501. 
A segmented person from a disjoint dataset is composited into query 
images at increasing coverage levels; gallery images remain clean. 
(left) Absolute mAP and Rank-1. (right) SAGA advantage over global 
\texttt{[CLS]}, peaking at \textbf{+10.1 Rank-1 pp} mean and 
\textbf{+6.8 mAP pp} mean for SAGA-Fusion near 60\% coverage. Shaded 
bands show $\pm$1 SD across distractor identities and entry angles.}
\label{fig:distractor}
\end{figure}

To evaluate SAGA's aggregation mechanism under a qualitatively 
harder occlusion condition than synthetic masking, we composite 
a segmented person from a disjoint dataset into each query image 
at coverage levels from 0\% to 80\%, with gallery images remaining 
clean throughout. Unlike masking, which removes identity signal, 
a human distractor introduces semantically confusing patches that 
compete with the target in anchor attention — a closer approximation 
to real deployment scenarios involving crowd occlusion.

Figure~\ref{fig:distractor} shows that the advantage pattern is 
consistent with the masking experiment but with two notable 
differences, both consistent with predictions two and three. First, 
the crossover point — where SAGA begins outperforming global 
\texttt{[CLS]} — is shifted to higher coverage, reflecting the 
additional difficulty of suppressing person patches relative to 
blank regions. Second, the SAGA-Refined mAP dip at low coverage 
is more pronounced than under masking: when the distractor is small 
and the query person largely intact, the distractor's semantically 
rich patches create stronger competition in anchor alignment than 
absent patches do, making reconstruction through a fixed basis 
noisier in this regime.

Both effects are mechanistically consistent: the anchor basis 
emerges from standard ReID supervision rather than explicit 
occlusion handling, so its advantage over global pooling grows 
only as corrupted patches begin to dominate. SAGA-Fusion remains 
positive throughout both conditions, with the global \texttt{[CLS]} 
component compensating in the low-coverage regime where 
reconstruction alone is at a disadvantage — confirming that fusion 
addresses a complementary weakness rather than merely adding signal.

\subsection{Comparison with State-of-the-Art}
\begin{table*}[h]
\centering
\caption{Comparison with state-of-the-art methods on standard and occluded ReID benchmarks. Methods are grouped by input modality and supervision setting. $^\dagger$Results reproduced using authors' released code and configurations; reproduced Market-1501 R1 differs from reported by 0.6\%, consistent with environment and random seed variance. $^\ddagger$No public implementation available; results are from the original paper and could not be reproduced under a consistent experimental protocol.}
\label{tab:main_results}
\setlength{\tabcolsep}{5pt}
\resizebox{\textwidth}{!}{
\begin{tabular}{l|l|cc|cc|cc|cc|cc|cc|cc}
\hline
\multirow{2}{*}{Method} 
& \multirow{2}{*}{Backbone} 
& \multicolumn{2}{c|}{Market-1501}
& \multicolumn{2}{c|}{DukeMTMC}
& \multicolumn{2}{c|}{MSMT17}
& \multicolumn{2}{c|}{Occluded-Duke} 
& \multicolumn{2}{c|}{P-DukeMTMC} 
& \multicolumn{2}{c|}{Occluded-ReID}
& \multicolumn{2}{c}{Occluded-Market}\\
\cline{3-16}
& & R1 & mAP & R1 & mAP & R1 & mAP & R1 & mAP & R1 & mAP & R1 & mAP & R1 & mAP\\
\hline
\multicolumn{16}{c}{\textbf{Image-based ReID Methods}} \\
\hline
PCB~\cite{sun2018pcb} & ResNet50
& 93.8 & 81.6 & 83.3 & 69.2 & -- & -- & -- & -- & -- & -- & 41.3 & 38.9 & 66.0 & 49.4\\
BOT~\cite{luo2019bag} & ResNet50
& 94.5 & 85.9 & 86.4 & 76.4 & -- & -- & -- & -- & -- & -- & -- & -- & -- & --\\
TransReID~\cite{he2021transreid} & ViT-B/16
& 95.2 & 88.9 & 90.7 & 82.0 & 85.3 & 67.4 & 66.4 & 59.2 & -- & -- & -- & -- & 78.2 & 64.7\\
CLIP-ReID~\cite{li2023clipreid} & ViT-B/16
& 95.4 & 90.5 & 90.8 & 83.1 & 89.7 & 75.8 & 67.2 & 60.3 & 79.5 & 68.7 & -- & -- & 79.5 & 68.7\\
PRO-FD~\cite{cui2024profd} & ViT-B/16
& 95.6 & 90.8 & 92.1 & 84.0 & -- & -- & 70.6 & 63.1 & 92.8 & 84.7 & 92.3 & 90.3 & -- & --\\
FCFormer~\cite{wang2024feature} & ViT-B/16
& -- & -- & -- & -- & -- & -- & 71.3 & 60.9 & 91.5 & 80.7 & 84.9 & 86.2 & -- & --\\
PADE~\cite{wang2024parallel} & ViT-B/16
& 95.8 & 89.8 & 91.3 & 82.8 & -- & -- & 72.3 & 63.0 & 89.3 & 84.8 & 83.7 & 79.9 & -- & --\\
SCING~\cite{xie2025scing} & ViT-B/16
& 96.2 & 91.0 & 91.3 & 83.7 & -- & -- & 71.1 & 63.4 & 93.7 & 84.4 & 93.8 & 90.9 & 80.3 & 69.2\\
PromptSG$^\ddagger$~\cite{yang2024pedestrian} & ViT-B/16
& 97.0 & 94.6 & 91.0 & 81.6 & 92.6 & 87.2 & -- & -- & -- & -- & -- & -- & -- & --\\
\hline
\multicolumn{16}{c}{\textbf{Video-based ReID Methods}} \\
\hline
CLIMB-ReID$^\dagger$~\cite{yu2025climb} & ViT-B/16
& 96.3 & 92.0 & -- & -- & 89.6 & 76.9 & 72.6 & 66.2 & -- & -- & 87.9 & 85.9 & 88.6 & 78.7\\
\hline
\multicolumn{16}{c}{\textbf{Methods with External Supervision}} \\
\hline
CLIP-SCGI$^\ddagger$~\cite{han2024clip} & ViT-B/16
& \textbf{97.6} & \textbf{96.0} & 91.3 & 81.6 & \textbf{92.9} & \textbf{88.2} & 67.5 & 59.2 & -- & -- & -- & -- & -- & --\\
\hline
\multicolumn{16}{c}{\textbf{Ours (Image-based, No External Supervision)}} \\
\hline
\textbf{SAGA-ReID (Ours)} & ViT-B/16
& 96.1 & 91.5
& \textbf{92.3} & \textbf{84.5}
& 90.5 & 77.4
& \textbf{77.8} & 68.3
& \textbf{94.4} & \textbf{85.7}
& \textbf{94.5} & \textbf{92.0}
& 88.6 & 76.0\\
\textbf{SAGA+CLIMB (Ours)} & ViT-B/16
& 96.6 & 92.7
& -- & --
& 90.0 & 78.4
& 77.7 & \textbf{70.8}
& -- & --
& 89.9 & 88.4
& \textbf{90.1} & \textbf{81.5}\\
\hline
\end{tabular}
}
\end{table*}
Our central claim is that the bottleneck in CLIP-based ReID lies in 
\emph{feature aggregation rather than representation capacity}. 
The controlled experiments in Sections~\ref{sec:occlusion} 
and~\ref{sec:distractor} isolate this mechanism directly; 
Table~\ref{tab:main_results} evaluates whether the same advantage 
holds at scale across standard benchmarks. Improvements should be 
most pronounced where global pooling is unreliable — under occlusion 
and cross-camera variation — rather than on clean benchmarks. The 
results consistently support this trend.

\paragraph{Robustness under occlusion.}
On Occluded-DukeMTMC, SAGA-ReID achieves \textbf{77.8/68.3} R1/mAP, 
outperforming CLIP-ReID by \textbf{+10.6/+8.0} and surpassing SCING 
(71.1/63.4), PRO-FD (70.6/63.1), and CLIMB-ReID (72.1/65.4) by clear 
margins. On Occluded-ReID and Occluded-Market, SAGA-ReID achieves 
\textbf{94.5/92.0} and \textbf{88.6/76.0} R1/mAP, outperforming all 
methods without external supervision or temporal information. These 
gains are consistent with the peak advantage observed in the 
controlled experiments, where structured reconstruction most benefits 
intermediate occlusion levels where partial identity evidence remains 
but global pooling increasingly conflates informative and corrupted 
regions.

\paragraph{Standard and cross-camera benchmarks.}
On Market-1501 and DukeMTMC-reID, SAGA-ReID achieves \textbf{96.1/91.5} 
and \textbf{92.3/84.5} R1/mAP, improving over CLIP-ReID on both. On 
MSMT17, gains of \textbf{+0.8/+1.6} R1/mAP are consistent with 
aggregation quality mattering under cross-camera shift, though more 
modest than under occlusion as expected. The smaller gains on clean 
benchmarks mirror the low-coverage regime of the controlled 
experiments, where global pooling remains competitive and the 
reconstruction constraint offers limited advantage.

\paragraph{Comparison with language-conditioned and dedicated 
aggregation methods.}
CLIP-SCGI incorporates LLaVA-generated captions as external supervision 
during training but lacks a reproducible implementation; we compare 
against reported numbers only. Despite no caption supervision, 
SAGA-ReID substantially outperforms CLIP-SCGI on Occluded-DukeMTMC 
(77.8 vs.\ 67.5 R1) and remains competitive on Market-1501 (96.1 
vs.\ 97.6 R1). PromptSG retains language at inference through an 
image-conditioned prompt but also lacks a public implementation; both 
methods' reported numbers are therefore unverified under consistent 
experimental conditions. Critically, as Figure~\ref{fig:methods_comparison} 
illustrates, both methods use language as a per-instance conditioning 
signal rather than a structured reconstruction basis — a distinction 
that the distractor experiment makes concrete: text-query-based 
suppression cannot distinguish target from distractor patches when 
both are person patches, whereas anchor-guided reconstruction 
down-weights the distractor through identity alignment rather than 
person-versus-background filtering.

CLIMB-ReID introduces a Mamba-based aggregation mechanism and achieves 
strong results on standard benchmarks, but as 
Section~\ref{sec:feature_analysis} shows, its Mamba feature 
underperforms the global \texttt{[CLS]} token on occluded single-image 
benchmarks — competitive results reflect improved backbone finetuning 
rather than effective sequential aggregation. This confirms prediction 
four: dedicated aggregation without a structured prior does not address 
the aggregation bottleneck regardless of architectural sophistication. 
SAGA-ReID outperforms CLIMB-ReID on R1 across all occluded benchmarks 
and on both metrics on Occluded-DukeMTMC and Occluded-ReID, despite 
starting from a weaker backbone — confirming that the aggregation 
mechanism contributes independently of backbone quality. The stronger 
CLIMB-ReID mAP on Occluded-Market is consistent with its better-trained 
backbone \texttt{[CLS]} token, as our feature analysis shows; applying 
SAGA aggregation to that backbone improves mAP further.

\subsection{Feature Representation Analysis}
\label{sec:feature_analysis}
\begin{table}[h]
\centering
\caption{Analysis of feature representations at different stages of 
the proposed pipeline, and comparison with CLIMB-ReID feature 
representations. Results show that (1) semantic refinement 
significantly improves robustness under occlusion, (2) combining 
global and refined features yields the best performance, and (3) 
SAGA's anchor-guided aggregation outperforms CLIMB-ReID's Mamba 
feature despite starting from a weaker backbone initialization. 
Bold indicates best result within each backbone group.}
\label{tab:feature_analysis}
\setlength{\tabcolsep}{5pt}
\begin{tabular}{l|cc|cc}
\hline
\multirow{2}{*}{Feature Type} 
& \multicolumn{2}{c|}{Market-1501} 
& \multicolumn{2}{c}{Occluded-Duke} \\
\cline{2-5}
& R1 & mAP & R1 & mAP \\
\hline
\multicolumn{5}{c}{\textit{CLIP-ReID backbone}} \\
\hline
CLIP global (\texttt{[CLS]}) 
& 95.6 & 90.5 & 69.4 & 61.4 \\
CLIP + bottleneck projection 
& 95.5 & 89.9 & 68.6 & 60.8 \\
SAGA feature
& 95.7 & 88.8 & 77.8 & 64.8 \\
Concatenated feature 
& 95.9 & 90.0 & 71.1 & 63.6 \\
SAGA fusion$^a$ 
& \textbf{96.1} & \textbf{91.5} & \textbf{77.8} & \textbf{68.3} \\
\hline
\hline
\multicolumn{5}{c}{\textit{CLIMB-ReID backbone$^\dagger$}} \\
\hline
CLIMB global (\texttt{[CLS]})
& 96.2 & 92.3 & 74.2 & 67.5 \\
CLIMB Mamba feature
& 95.8 & 90.2 & 67.8 & 59.4 \\
CLIMB combined (reported)
& 96.2 & 91.8 & 72.6 & 66.2 \\
SAGA feature
& 95.4 & 89.8 & 75.1 & 65.8\\
SAGA fusion$^b$
& \textbf{96.6} & \textbf{92.7} & \textbf{77.7} & \textbf{70.8} \\
\hline
\end{tabular}
\begin{tablenotes}
\small
\item $^a$Fusion weights $(w_r, w_i) = (2, 0.2)$.
\item $^b$Fusion weights $(w_r, w_i) = (0.5, 1.0)$. The lower 
weight on the refined feature reflects the stronger CLIMB-ReID 
backbone \texttt{[CLS]} token, which contributes more complementary 
signal than in the CLIP-ReID setting.
\item $^\dagger$CLIMB-ReID \texttt{[CLS]} and Mamba features 
extracted from intermediate representations using authors' released 
code. The Mamba feature underperforms the global \texttt{[CLS]} 
token on both benchmarks, indicating that sequential patch filtering 
does not improve over global pooling on single-image occluded ReID 
without temporal information.
\end{tablenotes}
\end{table}

\paragraph{Global vs.\ reconstructed features.}
The CLIP \texttt{[CLS]} feature degrades sharply under occlusion 
(69.4 $\rightarrow$ 77.8 R1 with anchor reconstruction on 
Occluded-DukeMTMC). Bottleneck adaptation yields negligible gains, 
confirming improvements arise from the aggregation mechanism rather 
than feature transformation. Weighted fusion achieves the best 
overall performance by combining robustness from reconstruction with 
the completeness of global representations — consistent with the 
controlled experiments, where SAGA-Fusion remains positive throughout 
both masking and distractor conditions by compensating for the 
low-coverage regime where reconstruction alone is at a disadvantage.

\paragraph{Aggregation vs.\ backbone quality.}
Despite Mamba aggregation being CLIMB-ReID's primary contribution, 
its Mamba feature substantially underperforms its own \texttt{[CLS]} 
token — competitive results reflect improved backbone finetuning, not 
effective sequential aggregation. This is precisely the failure mode 
identified in Section~\ref{sec:intro}: dedicated aggregation 
mechanisms without a structured prior do not address the aggregation 
bottleneck, regardless of architectural sophistication. SAGA's 
reconstructed feature outperforms CLIMB-ReID's Mamba feature despite 
starting from a weaker backbone, and applying SAGA to CLIMB-ReID's 
backbone improves performance further, confirming that aggregation 
structure and backbone quality are orthogonal and additive axes of 
improvement.

\paragraph{Fusion compatibility and backbone alignment.}
Despite SAGA's reconstructed feature being stronger in isolation on 
the CLIMB-ReID backbone, the fused SAGA+CLIMB result does not 
substantially exceed SAGA fusion with the CLIP-ReID backbone. This is 
consistent with the informational complementarity argument introduced 
in Section~\ref{sec:method}: CLIP-ReID's backbone \texttt{[CLS]} 
token, trained primarily for image--text alignment, emphasizes 
globally coherent identity signal but is less selective about 
spatially stable regions — leaving substantial complementary evidence 
for SAGA's anchor-guided reconstruction to contribute. CLIMB-ReID's 
more extensively finetuned backbone produces a stronger \texttt{[CLS]} 
token that already captures more locally discriminative evidence, 
reducing the informational gap that structured reconstruction can 
fill. This is reflected in the much lower optimal refined feature 
weight for SAGA+CLIMB versus SAGA with CLIP-ReID 
backbone  — the two components contribute more 
overlapping signal, limiting the ceiling on fusion gain. The result 
suggests that the complementarity benefit of structured reconstruction 
is largest when paired with a backbone whose global feature 
emphasizes holistic rather than spatially selective evidence.

\subsection{Anchor Attention Visualization}
\label{sec:visualization}

\begin{figure}[h]
\centering
\includegraphics[width=\linewidth]{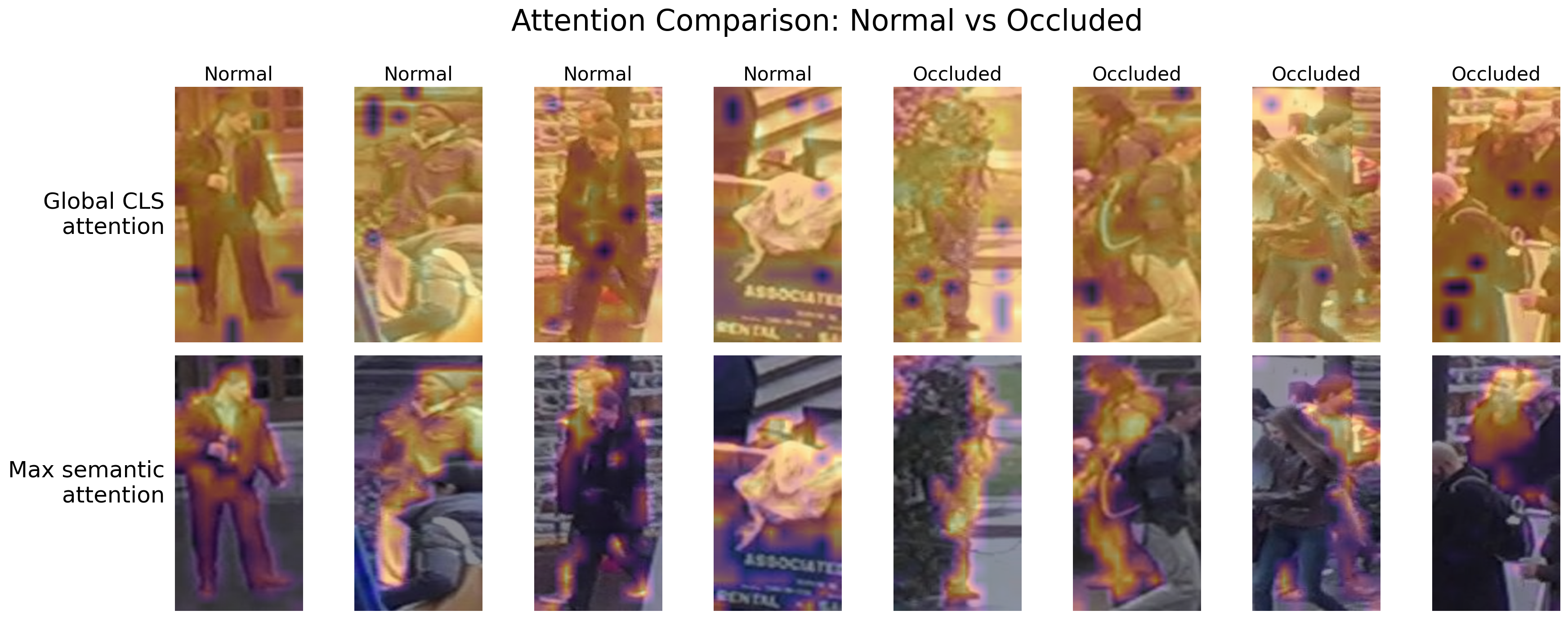}
\caption{Attention comparison on Occluded-DukeMTMC. Top: global CLS 
attention. Bottom: max structured-anchor attention. Structured-anchor 
attention remains person-centric under occlusion; CLS attention 
spreads across occluding objects.}
\label{fig:cls_vs_anchor}
\end{figure}

Figure~\ref{fig:cls_vs_anchor} compares global CLS and max-pooled 
structured-anchor attention. Structured-anchor attention concentrates 
on identity-relevant regions while suppressing occluders; CLS 
attention is more diffuse and frequently shifts toward interfering 
objects under occlusion.

\begin{figure}[h]
\centering
\includegraphics[width=\linewidth]{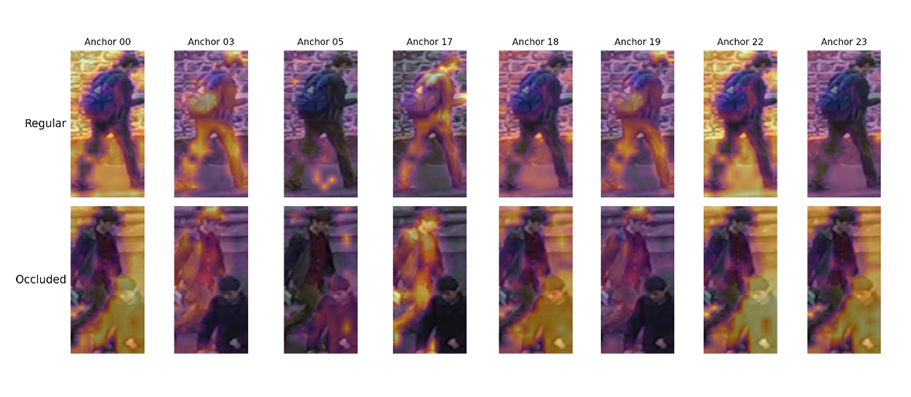}
\caption{Per-anchor attention maps for representative regular (top) 
and occluded (bottom) queries, showing the 8 maximally distinct 
anchors from the full set of 24. Anchors partition into distinct 
spatial behaviors rather than collapsing to a single pooling strategy. 
Anchors 03 and 19 show stronger lower-body activation in the regular 
case, becoming diffuse when occluded.}
\label{fig:per_anchor}
\end{figure}

Figure~\ref{fig:per_anchor} shows per-anchor attention maps. The 
anchor set partitions into distinct spatial behaviors — some 
activating broadly, others on localized regions. Anchors 03 and 19 
show the clearest spatial grounding, with lower-body activation that 
becomes diffuse when that region is occluded. The aggregate robustness 
of SAGA-ReID emerges from this structured division: when one region 
is corrupted, anchors focused on other regions continue contributing 
stable identity signal — arising purely from ReID supervision without 
pose annotations or part labels.

\subsection{Ablation Study}
\label{sec:ablation}
\begin{table}[h]
\centering
\caption{Ablation study on Occluded-Market. We compare free vs structure anchors and progressively add domain anchors and feature fusion.}
\label{tab:ablation_main}
\setlength{\tabcolsep}{6pt}
\begin{tabular}{l|c|c|c|c|cc}
\hline
Variant & FAR & SAR & DA & FF & R1 & mAP \\
\hline
\multicolumn{7}{c}{\textbf{Semantic anchor-based model}} \\
\hline
Baseline (CLIP-ReID)        &  &  &  &  & 67.2 & 60.3 \\
+ Structure anchors         &  & \checkmark &  &  & 77.4 & 64.1 \\
+ Domain anchors           &  & \checkmark & \checkmark &  & 77.8 & 64.8 \\
+ Feature fusion (final)   &  & \checkmark & \checkmark & \checkmark & \textbf{77.8} & \textbf{68.3} \\
\hline
\multicolumn{7}{c}{\textbf{Free anchor baseline (control)}} \\
\hline
+ Free anchors             & \checkmark &  &  &  & 76.7 & 63.9 \\
+ Free anchors + DA + FF   & \checkmark &  & \checkmark & \checkmark & 76.0 & 67.4 \\
\hline
\end{tabular}
\end{table}

Table~\ref{tab:ablation_main} progressively evaluates each component 
starting from a CLIP-ReID baseline. The primary gain comes from 
anchor-guided reconstruction itself: +10.2 R1 / +3.8 mAP over 
baseline, confirming \emph{aggregation structure} as the dominant 
factor. Replacing free anchors with structured anchors adds a 
consistent improvement (76.7 $\rightarrow$ 77.4 R1), demonstrating 
that text-space initialization contributes beyond raw cross-attention 
capacity. Domain anchors provide further improvement (77.8/64.8).

Feature fusion yields a larger mAP gain with structured anchors than 
with free anchors (64.8 $\rightarrow$ 68.3 vs.\ 63.9 $\rightarrow$ 
67.4), consistent with the informational complementarity argument 
introduced in Section~\ref{sec:method}. Text-space initialization 
combined with the decorrelation loss encourages anchors to partition 
into spatially distinct behaviors, producing a reconstructed feature 
that captures complementary evidence to the global \texttt{[CLS]} 
token. Free anchors, lacking this structured initialization, tend 
toward capturing globally salient regions already emphasized by 
\texttt{[CLS]}, contributing more redundant signal at fusion and 
limiting the mAP ceiling regardless of reconstruction quality.

\paragraph{Nature of text-space grounding.}
We augmented each anchor's context with a fixed text suffix 
(e.g., ``upper body of a person'') to test whether explicit semantic 
steering improves performance. It did not — performance was equivalent 
to fully learnable anchors and attention maps did not align with the 
specified concepts. We attribute this to operating at layer 
$\ell^* = 9$, where patch tokens have diverged from CLIP's joint 
embedding space. Text-space initialization therefore contributes 
through optimization structure and anchor diversity rather than 
enforced semantic correspondence — consistent with the per-anchor 
analysis in Section~\ref{sec:visualization}, which shows behaviorally 
distinct but not rigidly semantic patterns.

\subsection{Fusion Weight Sensitivity}
\begin{figure}[h]
\centering
\makebox[\linewidth][c]{%
\includegraphics[width=1.1\linewidth]{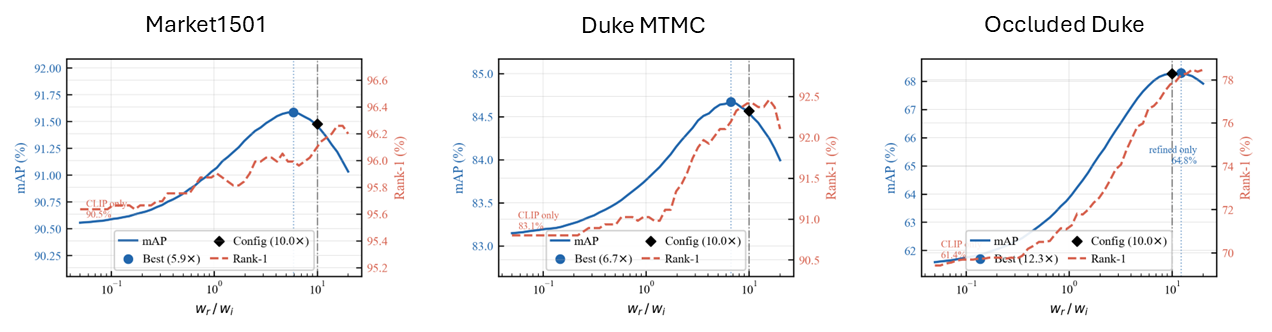}
}
\caption{\textbf{Effect of fusion weight ratio $w_r / w_i$ on 
retrieval performance.} mAP (left axis, blue) and Rank-1 (right 
axis, red dashed) as a function of $w_r / w_i$ with $w_r + w_i = 1$. 
Circle marks best mAP; diamond marks our fixed $10\times$ 
configuration.}
\label{fig:fusion_ratio}
\end{figure}
Performance rises sharply from the CLIP-only baseline ($w_r \to 0$), 
peaks at an intermediate ratio, and gradually declines as the backbone 
signal is suppressed — confirming complementarity between the two 
components. The larger optimal ratio on Occluded-DukeMTMC 
($12.3\times$ vs.\ $5.9\times$ on Market-1501) reflects the increased 
importance of structured reconstruction under occlusion. Our fixed 
$10\times$ configuration achieves near-optimal performance across all 
datasets without dataset-specific tuning.

\subsection{Computational Overhead}
SAGA-ReID adds $\approx$30M trainable parameters on top of the frozen 
ViT-B/16 backbone ($\approx$86M): Structured Aggregation Module 
($\approx$14.2M), Domain Anchor Generator ($\approx$14.8M), and 
projection heads ($\approx$1M). Inference overhead is negligible — 
$N{=}128$ patch tokens over $K{+}M{=}27$ anchors adds $\approx$5.3M 
FLOPs per image ($<$0.1\% of backbone cost). Stage 3 training 
completes in $\approx$38 minutes on Market-1501 on a single RTX 4080 
GPU with 16GB memory.

\subsection{Generalization to Unseen Domains}
\begin{table*}[h]
  \centering
  \small
  \resizebox{\linewidth}{!}{%
    \begin{tabular}{l|c|ccccccc}
      \hline
      \textbf{Method} & \textbf{Backbone} &
      $\rightarrow$clp & $\rightarrow$inf & $\rightarrow$pnt &
      $\rightarrow$qdr & $\rightarrow$rel & $\rightarrow$skt &
      \textbf{Avg.} \\
      \hline
      ERM++~\cite{teterwak2025erm++} & ViT-B/16 & -- & -- & -- & -- & -- & -- & 59.8 \\
      \hline
      CLIP (Prompt Tuning) & ViT-B/16 & 72.9 & 46.6 & 64.7 & 13.9 & 81.0 & 64.3 & 57.2 \\
      \textbf{SAGA (semantic anchors)} & ViT-B/16 & 77.4 & 50.3 & 70.2 & 17.5 & 82.8 & 68.0 & \textbf{61.0} \\
      \hline
    \end{tabular}
  }
  \caption{
    Classification accuracy (\%) on the DomainNet benchmark under the multi-source domain generalization task. ERM++ only reports average accuracy across all six target-domain settings.
  }
  \label{tab:domainnet_msdg}
\end{table*}
Table~\ref{tab:domainnet_msdg} evaluates SAGA's anchor mechanism in 
a multi-source domain generalization setting on 
DomainNet~\cite{peng2019moment}, training on five domains and 
evaluating on the held-out target. Replacing prompt tuning with 
structured anchor-guided aggregation improves average accuracy from 
57.2\% to 61.0\%, surpassing ERM++ ~\cite{teterwak2025erm++} (59.8\%) 
— a strong 2025 domain generalization baseline — with consistent 
gains across all six target domains. A full investigation including 
dedicated domain generalization baselines is left for future work.
\section{Limitations}
The structured anchors are learned under standard ReID supervision on a 
fixed training set, and while the domain anchor module provides 
instance-level adaptation, the shared anchor representations are static 
at inference. This may limit robustness when deployed across 
substantially different camera networks or imaging conditions not 
represented during training. Lightweight anchor adaptation at deployment 
time — for example via few-shot or test-time adaptation of the context 
vectors — is a natural direction for future work.

The claim that feature aggregation is the primary bottleneck in 
CLIP-based ReID is supported through specific instantiations — SAGA-ReID 
applied to a ViT-B/16 backbone with CLIP-ReID initialization, and 
extended to CLIMB-ReID features. Whether this finding generalizes to 
other CLIP-based architectures or pretraining strategies remains an open 
question.
\section{Conclusion}
We have argued that feature aggregation is a primary bottleneck in 
CLIP-based person re-identification — one that persists even in methods 
that introduce dedicated aggregation mechanisms. Analysis of CLIMB-ReID 
shows its Mamba feature underperforms the global \texttt{[CLS]} token 
on single-image occluded benchmarks, confirming the problem requires a 
structured prior rather than architectural complexity alone.

SAGA-ReID addresses this with a targeted intervention: anchor vectors 
parameterized in CLIP's text embedding space serve as a structured basis 
for patch feature reconstruction via cross-attention. Language serves 
not as supervision but as a \emph{structural prior} that governs how 
spatial evidence is composed at inference, without requiring per-image 
textual descriptions or part annotations.

Controlled experiments under synthetic masking and human distractor 
occlusion validate the mechanism directly, with SAGA's advantage peaking 
at \textbf{+30 Rank-1 pp} and \textbf{+10.1 Rank-1 pp} respectively — 
the peak-then-decline pattern consistent across both conditions confirms 
structured reconstruction is most beneficial where partial identity 
evidence remains but global pooling increasingly conflates informative 
and corrupted regions. Benchmark evaluations confirm consistent gains 
over CLIP-ReID with the largest improvements under occlusion, and 
SAGA's reconstructed feature outperforms CLIMB-ReID's Mamba feature 
despite starting from a weaker backbone. Structured anchor aggregation 
also improves multi-source domain generalization on DomainNet over both 
prompt tuning and a strong 2025 baseline, suggesting applicability 
beyond ReID.

\paragraph{Broader Impacts.}
Person re-identification has direct applications in multi-camera 
surveillance and pedestrian tracking, which carry inherent privacy 
implications. The primary contribution here is a feature aggregation 
method evaluated on standard academic benchmarks; no deployed system 
or novel data collection is introduced. Researchers and practitioners 
applying this work in real systems should consider applicable privacy 
regulations and consent frameworks.
\bibliographystyle{unsrtnat}
\bibliography{neurips_2026}



\newpage

\end{document}